\documentclass[runningheads]{llncs}
\usepackage{cite}
\usepackage{amsmath,amssymb,amsfonts}
\usepackage{algorithmic}
\usepackage{graphicx}
\usepackage{textcomp}
\usepackage{xcolor}
\usepackage{booktabs}
\usepackage{enumitem}
\usepackage[colorlinks=true, linkcolor=blue, urlcolor=blue]{hyperref}
\usepackage{multirow}
\usepackage{tabularx}
\usepackage{longtable}
\usepackage{array} 
\usepackage{booktabs} 
\usepackage{adjustbox} 
\usepackage[T1]{fontenc}
\usepackage{graphicx}
\usepackage{xparse}

\makeatletter
\DeclareDocumentCommand\todo{g}{%
\def\@message{\IfNoValueTF{#1}{TODO}{TODO: #1}}
\textbf{\textcolor[HTML]{FF8811}{\@message}}
\@latex@warning{\@message}{}{}}
\makeatother
\begin{document}
%

\title{Automated Bias Assessment in AI-Generated Educational Content Using CEAT Framework}

\titlerunning{Automated Bias Assessment in AI-Generated Educational Content}

\author{Jingyang Peng\inst{1} \and
Wenyuan Shen \inst{1} \and
Jiarui Rao\inst{1} \and 
Jionghao Lin\inst{2, 1, }\thanks{Corresponding author.}}
\authorrunning{J. Peng et al.}
%
\institute{
    Carnegie Mellon University, Pittsburgh PA 15213, USA \\
    \email{{\{jingyanp, wenyuan2, jiaruira, jionghal\}@andrew.cmu.edu }} \and
    The University of Hong Kong, Pokfulam Rd, Hong Kong, China\\
    \email{jionghao@hku.hk} 
}

\maketitle              

\begin{abstract} 

Recent advances in Generative Artificial Intelligence (GenAI) have transformed educational content creation, particularly in developing tutor training materials. However, biases embedded in AI-generated content—such as gender, racial, or national stereotypes—raise significant ethical and educational concerns. Despite the growing use of GenAI, systematic methods for detecting and evaluating such biases in educational materials remain limited. This study proposes an automated bias assessment approach that integrates the Contextualized Embedding Association Test (CEAT) with a prompt-engineered word extraction method within a Retrieval-Augmented Generation (RAG) framework. We applied this method to AI-generated texts used in tutor training lessons. Results show a high alignment between the automated and manually curated word sets, with a Pearson correlation coefficient of \(r = 0.993\), indicating reliable and consistent bias assessment. Our method reduces human subjectivity and enhances fairness, scalability, and reproducibility in auditing GenAI-produced educational content.

 
\end{abstract}

\section{Background}

Recent advancements in Generative Artificial Intelligence (GenAI), particularly Large Language Models (LLMs), have led to the development of systems aimed at assisting with educational content creation and personalized learning, demonstrating promising potential for enhancing teaching and learning \cite{lin2024can,moore2023empowering,dai2024assessing}. However, biases embedded within AI-generated content raise substantial ethical and societal concerns, as these biases typically originate from biased or unrepresentative training datasets~\cite{Ferrara-GPT,Survey_on_Fairness}. Specifically, demographic biases, which reflect societal inequalities related to characteristics such as gender, race, nationality, and ethnicity, often manifest in GenAI output by reinforcing harmful stereotypes. For example, generative AI has been shown to associate Muslims with terrorism~\cite{Muslim} or incorrectly link Black individuals with crime~\cite{Ask_LLMs_Directly}. Such biases compromise educational equity and further risk perpetuating existing social prejudices~\cite{Ferrara-Survey}.

To systematically detect these biases, computational approaches like the Word Embedding Association Test (WEAT) have been widely employed, measuring associations between predefined sets of target words (e.g., demographic groups such as ``\textit{male}'' and ``\textit{female}'') and attribute words (e.g., descriptive terms such as ``\textit{career-oriented}'' and ``\textit{family-oriented}'')~\cite{WEAT}. Despite its effectiveness, WEAT evaluates words in isolation without considering how meanings shift based on context, significantly limiting its ability to detect biases in realistic, context-rich educational materials. To overcome this limitation, the Contextualized Embedding Association Test (CEAT) was introduced as an extension of WEAT, incorporating contextual information by assessing word associations within actual sentences and discourses~\cite{CEAT}. Although CEAT improves the ability to detect biases as they naturally occur in context, it still relies on manually curated word sets, introducing potential subjectivity and inefficiency in bias evaluations. To overcome the challenges associated with the manual selection of target and attribute word sets, we propose an automated method for generating context-sensitive word sets for bias assessment. Specifically, our study addresses the research question: \textit{How does the automated extraction method compare to manually defined ground-truth word sets in CEAT-based bias evaluations?}

\vspace{-1mm}
\section{Method}

\subsection{Data}


We used a dataset consisting of 10 AI-generated lesson scripts developed for online tutor training, many of which exhibited demographic biases (e.g., gender, racial, and national). These texts were systematically analyzed using a structured annotation process to construct ground-truth rubrics for bias identification. To evaluate the performance of our word extraction method, we also used a separate set of 4 educational texts. Full dataset descriptions, annotation protocols, and the code for automated extraction are available at \href{https://github.com/EricP66/Automated-Word-Extraction}{Github Repository}.

\vspace{-2mm}





\subsection{Bias Evaluation using CEAT}

We applied the Contextualized Embedding Association Test (CEAT)~\cite{CEAT} to quantify bias within contextualized embeddings. Unlike static embedding tests such as WEAT~\cite{WEAT}, CEAT captures bias by considering word associations within specific textual contexts. The CEAT approach involves the following key steps:

\paragraph{Step 1: Computing Word Association Scores.}  
We define two equal-sized sets of \textit{target words} \(X\) and \(Y\), each representing specific demographic groups (e.g. gender or nationality), and two equal-sized sets of \textit{attribute words} \(A\) and \(B\), which describe characteristics or qualities associated with the target groups. For each target word \(w\), we calculate its association score as follows:

\begin{equation}
s(\vec{w}, A, B) = \text{mean}_{a \in A}\cos(\vec{w}, \vec{a}) - \text{mean}_{b \in B}\cos(\vec{w}, \vec{b})
\end{equation}

Here, \(\cos(\vec{w}, \vec{a})\) represents the cosine similarity between the embedding of the word \(w\) and each attribute word \(a\), and \(\text{mean}_{a \in A}\cos(\vec{w}, \vec{a})\) stands for the average cosine similarity between the target word \(w\) and all attribute words in word set \(A\). The difference in average cosine similarity reflects how strongly a target word \(w\) is associated with two attribute word sets, with a positive score indicating a stronger association with attributes in \(A\) and a negative score suggesting a stronger association with attributes in \(B\).
\vspace{1mm}

\noindent\textit{Step 2: Calculating Effect Size (ES).}  
The CEAT method quantifies the relative bias between two sets of target words (\(X\) and \(Y\)) using the calculated association scores. The Effect Size (ES) reflects how strongly one demographic group is associated with a particular attribute set compared to another group:

\begin{equation}
ES(X, Y, A, B) = \frac{\text{mean}_{x \in X}s(x, A, B) - \text{mean}_{y \in Y}s(y, A, B)}{\text{std\_dev}_{w \in X \cup Y}s(w, A, B)}
\end{equation}

The numerator in Equation (2) captures the difference in the average association scores between two set of targets, and the denominator normalizes this difference using the standard deviation of the scores for all target words. A higher absolute value of ES indicates a stronger bias between the two demographic groups.
\vspace{1mm}

\noindent\textit{Step 3: Computing Combined Effect Size (CES).}  
To summarize the bias measurements across multiple contexts, CEAT applies a random-effects model, which produces a Combined Effect Size (CES). CES is calculated as the weighted average of all individual effect sizes (\(ES_i\)) in \(N\) different contextual samples:

\begin{equation}
CES(X, Y, A, B) = \frac{\sum_{i=1}^N v_i \cdot ES_i}{\sum_{i=1}^N v_i},\quad \text{where}\quad v_i = \frac{1}{\sigma^2_{\text{within}} + \sigma^2_{\text{between}}}
\end{equation}

Here, \(v_i\) denotes the inverse of the total variance associated with each context \(i\), considering both within-sample variance (\(\sigma^2_{\text{within}}\)) and between-sample variance (\(\sigma^2_{\text{between}}\)). The combination of various effect sizes offers a more robust measurement of bias in diverse contexts.
\vspace{1mm}

\noindent\textit{Step 4: Interpreting Magnitude of Bias.}  
Finally, we interpret the resulting CES scores using Cohen’s \(d\) benchmarks~\cite{Rice_Harris}. Values of 0.2, 0.5, and 0.8 indicate small, medium and large effect sizes respectively, helping contextualize the magnitude and potential impact of identified biases.





\subsection{Automated Word Extraction}

Manual creation of target and attribute sets is inefficient and subjective~\cite{CEAT}. Therefore, we developed an automated extraction method that leverages prompt engineering within a Retrieval-Augmented Generation (RAG) framework~\cite{RAG}.

\noindent\textbf{Text Pre-processing.}
We split input texts into contextually coherent chunks and generated OpenAI embeddings for efficient retrieval and generation.

\noindent\textbf{Few-Shot Prompting.} 
We implemented few-shot prompting with \textit{GPT-4o} to guide the automated extraction of demographic biases. Specifically, we designed prompts that provide clear task definitions and include illustrative examples for identifying common forms of demographic bias—namely gender, national, and racial biases. Full prompt details, including templates and constraints, are available at \href{https://github.com/EricP66/Automated-Word-Extraction}{Github Repository}.

\noindent\textbf{Extraction Goals and Constraints.}  
Prompts included explicit goals (accurate identification of demographic groups) and constraints (no inferred terms, exhaustive extraction, accurate representation of text formatting). If multiple demographic groups were identified, we computed averaged pairwise CEAT scores to ensure balanced representation.

\noindent\textbf{Rubrics and Annotation Procedure.}  
To evaluate the performance of our automated method, we created ground truth word sets through manual annotation. Three trained annotators independently reviewed each AI-generated text and extracted target and attribute word sets based on the presence of explicit or implicit demographic indicators. The annotation process followed detailed rubrics (introduced in the \href{https://github.com/EricP66/Automated-Word-Extraction}{GitHub Repository}), which ensured systematic and unbiased identification of demographic group terms and associated attributes. Any discrepancies among annotators were resolved through discussion to reach consensus. This structured annotation workflow minimized subjectivity and enabled a fair, quantitative comparison with the extracted word sets.

\section{Results}

\subsection{Automated Word Set Extraction}


To assess the performance of our automated extraction method, we systematically compared automatically generated target and attribute word sets against manually curated ground-truth sets across multiple AI-generated educational texts. As a representative example, we conducted an in-depth comparative analysis on the AI-generated course titled \textit{``Profiles of Student Success: Inspiring Stories Across Diverse Backgrounds,''} which discusses student achievements from various cultural and demographic groups. Ground-truth word sets for this course were derived through our predefined rubric-based procedure, while the automated word sets were generated using our RAG-based method with carefully designed prompt engineering. Table~\ref{tab:comparison_ground_truth_automated_adjusted} illustrates the comparison of these word sets, along with computed cosine similarity scores to quantitatively measure the semantic alignment between automated and ground-truth extractions.

\begin{table}[ht]
\vspace{-3mm}
\centering
\renewcommand{\arraystretch}{1.1}
\caption{Comparison of Ground Truth and Automated Extraction Word Sets}
\label{tab:comparison_ground_truth_automated_adjusted}
\resizebox{1\textwidth}{!}{%
\begin{tabular}{|l|l|l|l|l|c|}
\hline
\textbf{\begin{tabular}[c]{@{}l@{}}Demographic\\ Group\end{tabular}} & \textbf{\begin{tabular}[c]{@{}l@{}}Target Words \\ (Ground Truth)\end{tabular}}   & \textbf{\begin{tabular}[c]{@{}l@{}}Target Words \\ (Extracted)\end{tabular}} & \textbf{\begin{tabular}[c]{@{}l@{}}Attribute Words \\ (Ground Truth)\end{tabular}}                                             & \textbf{\begin{tabular}[c]{@{}l@{}}Attribute Words\\ (Extracted)\end{tabular}}  &
\textbf{\begin{tabular}[c]{@{}l@{}} Similarity \\ Scores \end{tabular}}   \\ \hline
Mexican                    & \begin{tabular}[c]{@{}l@{}}Mexico\\ Mexican\\ Carlos Ramirez\end{tabular}         & \begin{tabular}[c]{@{}l@{}}Mexican\\ Mexico\\ Carlos Ramirez\end{tabular}               & \begin{tabular}[c]{@{}l@{}}talent\\ perseverance \\ discipline\\ analytical\\ structured environment\end{tabular}              & \begin{tabular}[c]{@{}l@{}}talent\\ discipline\\ perseverance\\ analytical mind\\ natural ability\end{tabular}  &
 \begin{tabular}[c]{@{}l@{}} 0.7628 \end{tabular} \\ \hline
American                   & \begin{tabular}[c]{@{}l@{}}United States\\ American\\ Sarah Thompson\end{tabular} & \begin{tabular}[c]{@{}l@{}}American\\ United States\\ Sarah Thompson\end{tabular}       & \begin{tabular}[c]{@{}l@{}}independence\\ empathy\\ advocacy\\ creativity\\ leadership\\ social consciousness\end{tabular}     & \begin{tabular}[c]{@{}l@{}}independence\\ creativity\\ entrepreneurial\\ individualism\\ social consciousness\end{tabular}          
&
 \begin{tabular}[c]{@{}l@{}} 0.7833\end{tabular} \\\hline
Chinese                    & \begin{tabular}[c]{@{}l@{}}China\\ Chinese\\ Li Wei\end{tabular}                  & \begin{tabular}[c]{@{}l@{}}Chinese\\ China\\ Li Wei\end{tabular}                        & \begin{tabular}[c]{@{}l@{}}precision\\ commitment\\ academic achievement\\ discipline\\ methodical\end{tabular}                & \begin{tabular}[c]{@{}l@{}}discipline\\ academic achievement\\ precision\\ commitment\\ long-term goals\end{tabular}  &
 \begin{tabular}[c]{@{}l@{}} 0.8216\end{tabular}              \\ \hline
Nigerian                   & \begin{tabular}[c]{@{}l@{}}Nigeria\\ Nigerian\\ Aisha Mohammed\end{tabular}       & \begin{tabular}[c]{@{}l@{}}Nigerian\\ Nigeria\\ Aisha Mohammed\end{tabular}             & \begin{tabular}[c]{@{}l@{}}resilience\\ creativity\\ empowerment\\ innovation\\ resourcefulness\end{tabular}                   & \begin{tabular}[c]{@{}l@{}}creativity\\ resilience\\ empowerment\\ breaking traditional norms\\ innovation\end{tabular}  &
 \begin{tabular}[c]{@{}l@{}} 0.8040 \end{tabular}                       \\ \hline
German                     & \begin{tabular}[c]{@{}l@{}}Germany\\ German\\ Michael Jensen\end{tabular}         & \begin{tabular}[c]{@{}l@{}}German\\ Germany\\ Michael Jensen\end{tabular}               & \begin{tabular}[c]{@{}l@{}}practicality\\ technical expertise\\ hands-on\\ vocational training\\ applied learning\end{tabular} & \begin{tabular}[c]{@{}l@{}}hands-on\\ practical skills\\ technical expertise\\ applied learning\\ vocational training\end{tabular} &
 \begin{tabular}[c]{@{}l@{}} 0.8895 \end{tabular}             \\ \hline
Indian-American            & \begin{tabular}[c]{@{}l@{}}Indian-American\\ Indian\\ Priya Patel\end{tabular}    & \begin{tabular}[c]{@{}l@{}}Indian\\ Indian-American\\ Priya Patel\end{tabular}          & \begin{tabular}[c]{@{}l@{}}cultural heritage\\ innovation\\ holistic\\ responsibility\\ adaptability\end{tabular}              & \begin{tabular}[c]{@{}l@{}}tradition\\ innovation\\ cultural heritage\\ modern medical practices\\ holistic approaches\end{tabular} &
 \begin{tabular}[c]{@{}l@{}} 0.7627\end{tabular}            \\ \hline
\end{tabular}
}
\vspace{-3mm}
\end{table}

According to Table~\ref{tab:comparison_ground_truth_automated_adjusted}, automated word sets show a strong alignment with the manually curated ground-truth sets. Target word sets from both approaches matched exactly, while attribute word sets exhibited significant overlap across all demographic groups. Cosine similarity scores ranged between 0.7627 and 0.8895, surpassing the 0.7 threshold indicative of strong semantic similarity~\cite{Cosine-proof}. Although minor differences exist, the consistently high similarity underscores the accuracy and reliability of our automated extraction method.

\subsection{CEAT-based Evaluation of Automated Extraction}

To further validate our method, we compared CEAT scores computed using automated and ground-truth word sets from multiple AI-generated educational texts (Table~\ref{tab:comparison_text_score}). Results indicate minimal deviations between automated and ground-truth-derived CEAT scores. The highest correlation was seen in Course~3 (0.2301 vs.~0.2406), while the largest deviation occurred in Course~1 (-0.1274 vs.~-0.1014), still within an acceptable margin. These results suggest that our automated extraction method effectively captures bias in AI-generated texts.


\begin{table}[h!]
\centering
\scriptsize
\caption{CEAT Score of Ground Truth and Automated Extraction Word Sets}
\label{tab:comparison_text_score}
\renewcommand{\arraystretch}{1.45}
\resizebox{0.8\textwidth}{!}{%
\begin{tabular}{|p{3cm}|>{\centering\arraybackslash}p{3cm}|>{\centering\arraybackslash}p{3.5cm}|}
\hline
\textbf{Text Score} & \textbf{Ground Truth} & \textbf{Automated Extraction} \\ \hline
Tutorial Course 1  & -0.1273  & -0.1014 \\ \hline
Tutorial Course 2  & 0.0428  & 0.0191 \\ \hline
Tutorial Course 3  & 0.2300 & 0.2406 \\ \hline
Tutorial Course 4  & -0.1664  & -0.1721 \\ \hline
\end{tabular}
}
\vspace{-3mm}
\end{table}


To rigorously verify the alignment between automated and ground-truth CEAT scores, we computed the Pearson correlation coefficient, which measures the strength of linear relationships between two variables~\cite{Pearson}. Our results revealed an exceptionally high correlation of \( r = 0.9930 \), indicating a near-perfect positive alignment between the two approaches. Such a strong correlation underscores that our automated extraction method reliably replicates ground-truth bias evaluations, confirming its effectiveness as a scalable and objective tool for bias assessment in AI-generated educational content.

\section{Discussion and Conclusion}

This study introduced an automated word extraction approach, leveraging prompt engineering within a Retrieval-Augmented Generation (RAG) framework, to systematically and objectively assess demographic biases in AI-generated educational content using GPT-4o model. The proposed method was validated through a strong Pearson correlation (\( r = 0.9930 \)) between automated and manually curated CEAT bias evaluation scores, confirming its potential as a scalable and reliable bias assessment tool.
\vspace{2mm}

\noindent\textbf{Educational Implications.} Our approach provides an objective method for evaluating biases embedded in AI-generated educational materials, such as digital textbooks, automated learning modules, or AI-powered tutoring platforms. By proactively identifying bias, educators can enhance teaching practices by ensuring educational content is inclusive and equitable for diverse student populations. Furthermore, this method can inform policy-related initiatives aimed at regulating and monitoring the fairness and ethical standards of GenAI tools employed in educational settings.
\vspace{2mm}

\noindent\textbf{Limitations and Future Work.} Despite its promising results, our study has several limitations. First, our validation currently relies on a limited dataset of AI-generated and human-crafted educational texts; broader validation across various educational contexts and larger-scale case studies in actual classroom settings are necessary to further validate the practical effectiveness and generalizability of our approach. Second, while identifying biases is essential, effectively mitigating them is equally crucial. Our work focuses exclusively on bias detection and does not address mitigation. In the future, we should explore bias mitigation strategies across different model training phases—such as counterfactual data augmentation during pre-training~\cite{CDA,Ferrara-Survey}, adversarial training or selective parameter freezing during model training~\cite{Adversarial_Learning,Bias_and_Fairness_in_Large_Language_Models}, and human-in-the-loop approaches in the post-training phase~\cite{Bias_and_Fairness_in_Large_Language_Models}. 
\vspace{2mm}

\section*{Acknowledgment} We sincerely thank Professor Kenneth R. Koedinger for his valuable discussions and insightful suggestions on this work. 

This work was fully supported by a grant from the University Research Committee (Grant No. 2401102970) at The University of Hong Kong. The opinions, findings, and conclusions expressed in this paper are solely those of the authors.


%
%
%
\bibliographystyle{plain}
\bibliography{mybibliography}

\begin{thebibliography}{10}

\bibitem{Muslim}
Abubakar Abid, Maheen Farooqi, and James Zou.
\newblock Persistent anti-muslim bias in large language models, 2021.

\bibitem{WEAT}
Aylin Caliskan, Joanna~J. Bryson, and Arvind Narayanan.
\newblock Semantics derived automatically from language corpora contain human-like biases.
\newblock {\em Science}, 356(6334):183–186, April 2017.

\bibitem{Pearson}
Israel Cohen, Yiteng Huang, Jingdong Chen, Jacob Benesty, Jacob Benesty, Jingdong Chen, Yiteng Huang, and Israel Cohen.
\newblock Pearson correlation coefficient.
\newblock {\em Noise reduction in speech processing}, pages 1--4, 2009.

\bibitem{dai2024assessing}
Wei Dai, Yi-Shan Tsai, Jionghao Lin, Ahmad Aldino, Hua Jin, Tongguang Li, Dragan Ga{\v{s}}evi{\'c}, and Guanliang Chen.
\newblock Assessing the proficiency of large language models in automatic feedback generation: An evaluation study.
\newblock {\em Computers and Education: Artificial Intelligence}, 7:100299, 2024.

\bibitem{Ferrara-GPT}
Emilio Ferrara.
\newblock Should chatgpt be biased? challenges and risks of bias in large language models.
\newblock {\em First Monday}, November 2023.

\bibitem{Ferrara-Survey}
Emilio Ferrara.
\newblock Fairness and bias in artificial intelligence: A brief survey of sources, impacts, and mitigation strategies.
\newblock {\em Sci}, 6(1), 2024.

\bibitem{Bias_and_Fairness_in_Large_Language_Models}
Isabel~O. Gallegos, Ryan~A. Rossi, Joe Barrow, Md~Mehrab Tanjim, Sungchul Kim, Franck Dernoncourt, Tong Yu, Ruiyi Zhang, and Nesreen~K. Ahmed.
\newblock Bias and fairness in large language models: A survey, 2024.

\bibitem{Adversarial_Learning}
Ian~J. Goodfellow, Jonathon Shlens, and Christian Szegedy.
\newblock Explaining and harnessing adversarial examples, 2015.

\bibitem{CEAT}
Wei Guo and Aylin Caliskan.
\newblock Detecting emergent intersectional biases: Contextualized word embeddings contain a distribution of human-like biases.
\newblock In {\em Proceedings of the 2021 AAAI/ACM Conference on AI, Ethics, and Society}, AIES ’21, page 122–133. ACM, July 2021.

\bibitem{RAG}
Patrick Lewis, Ethan Perez, Aleksandra Piktus, Fabio Petroni, Vladimir Karpukhin, Naman Goyal, Heinrich K{\"u}ttler, Mike Lewis, Wen-tau Yih, Tim Rockt{\"a}schel, et~al.
\newblock Retrieval-augmented generation for knowledge-intensive nlp tasks.
\newblock {\em Advances in Neural Information Processing Systems}, 33:9459--9474, 2020.

\bibitem{Survey_on_Fairness}
Yingji Li, Mengnan Du, Rui Song, Xin Wang, and Ying Wang.
\newblock A survey on fairness in large language models, 2024.

\bibitem{lin2024can}
Jionghao Lin, Zifei Han, Danielle~R Thomas, Ashish Gurung, Shivang Gupta, Vincent Aleven, and Kenneth~R Koedinger.
\newblock How can i get it right? using gpt to rephrase incorrect trainee responses.
\newblock {\em International journal of artificial intelligence in education}, pages 1--27, 2024.

\bibitem{Cosine-proof}
Tomas Mikolov.
\newblock Efficient estimation of word representations in vector space.
\newblock {\em arXiv preprint arXiv:1301.3781}, 3781, 2013.

\bibitem{moore2023empowering}
Steven Moore, Richard Tong, Anjali Singh, Zitao Liu, Xiangen Hu, Yu~Lu, Joleen Liang, Chen Cao, Hassan Khosravi, Paul Denny, et~al.
\newblock Empowering education with llms-the next-gen interface and content generation.
\newblock In {\em International Conference on Artificial Intelligence in Education}, pages 32--37. Springer, 2023.

\bibitem{Rice_Harris}
M.~E. Rice and G.~T. Harris.
\newblock Comparing effect sizes in follow-up studies: {ROC} area, cohen's d, and r.
\newblock {\em Law and Human Behavior}, 29:615--620, 2005.

\bibitem{Ask_LLMs_Directly}
Jisu Shin, Hoyun Song, Huije Lee, Soyeong Jeong, and Jong~C. Park.
\newblock Ask llms directly, "what shapes your bias?": Measuring social bias in large language models, 2024.

\bibitem{CDA}
Ran Zmigrod, Sabrina~J. Mielke, Hanna~M. Wallach, and Ryan Cotterell.
\newblock Counterfactual data augmentation for mitigating gender stereotypes in languages with rich morphology.
\newblock {\em CoRR}, abs/1906.04571, 2019.

\end{thebibliography}

\end{document}